\documentclass[a4paper,twoside]{article}
\usepackage{epsfig}
\usepackage{subfigure}
\usepackage{calc}
\usepackage{amssymb}
\usepackage{amstext}
\usepackage{amsmath}
\usepackage{amsthm}
\usepackage{multicol}
\usepackage{multirow}
\usepackage{pslatex}
\usepackage{apalike}
\usepackage{SciTePress}
\usepackage[small]{caption}
\usepackage{setspace}
\begin{document}

\title{\uppercase{Convergence Properties of Two $(\mu+\lambda)$ Evolutionary Algorithms on OneMax and Royal Roads Test Functions}}  

\author{\authorname{Aram Ter-Sarkisov\sup{1}, Stephen Marsland\sup{2}}
\affiliation{\sup{1}School of Computer Science, Massey University, Wellington, New Zealand}
\affiliation{\sup{2}School of Computer Science, Massey University, Palmerston North, New Zealand}
\email{\{a.ter-sarkisov, s.r.marsland\}@massey.ac.nz}
}


\abstract{We present a number of bounds on convergence time for two elitist population-based Evolutionary Algorithms using a recombination operator k-Bit-Swap and a mainstream Randomized Local Search algorithm. We study the effect of distribution of elite species and population size.}

\onecolumn \maketitle \normalsize \vfill

\section{\uppercase{Introduction}}
\label{sec:introduction}

\noindent The main objective of this article is to derive convergence properties of two elitist Evolutionary Algorithms (EAs) on OneMax and Royal Roads test functions. One of the analyzed algorithms uses the k-Bit-Swap (kBS) operator introduced in \cite{tersarkisov2010}. We compare our results to computational findings and other research.\\
\linebreak
A population consists of a set of solution strings. We split them into two groups: there are elite strings, which have the same highest fitness value and the remaining  non-elite strings. We use the standard notation for the population $\mu$, recombination pool $\lambda$, the elite species $\alpha$, the non-elite $\beta$.\\
\subsection{Past work}
Recently $(1+1)$EA with $\frac{1}{n}$ flip probability ($n$ being the length of a chromosome) became a matter of extensive investigation. Sharp lower and upper bounds for OneMax and general linear functions were found in \cite{doerr10,doerr102,doerr11,droste02} applying drift analysis and potential functions. Specifically, in \cite{doerr11} the upper bound for $(1+1)$EA solving OneMax was derived to be $(1+o(1))1.39en \log n$ and in \cite{doerr102} the lower bound for the same setting was found to be $(1-o(1))en\log n$. Drift (a form of super martingale) was introduced in \cite{hajek82,heyao03,heyao04}.\\
\linebreak
\linebreak
\subsection{Definitions and Assumptions}
We analyze two fitness functions here, OneMax (simple counting 1's test function) and Royal Roads (see Section \ref{sec:rr} for additional definitions for it). The fitness of a population is defined as the fitness of an elite string. Since both functions have global solution at $n$, we are interested in the following time parameter: 
{\smalll
\begin{equation}
\mathbf{\tau}_A=\min\{t \geq 0:f(\alpha)=n\}
\end{equation}   
}

\noindent that is, the minimum time when (for algorithm A) the best species in the population reaches the highest fitness value. Since the analysis is probabilistic, we need the expectation of this parameter: $ \mathbf{E}\tau_A$.\\
\linebreak
We assume that we do not need a large number of species for evolution. Though this sounds a bit vague, this justifies the choice of distributions with respective parameters. The expectation of Poisson random variable used here is 1, for Uniform it is $\frac{\mu+1}{2}$. We use the latter due to its simplicity.\\
\linebreak
We restrict our attention only to elite pairs (kBS) or parents (RLS), to simplify the analysis, since otherwise we would have to make more assumptions about the fitness of non-elite parents $\beta$.\\
\subsection{k-Bit-Swap Operator}
\begin{figure}
\includegraphics[scale=.35]{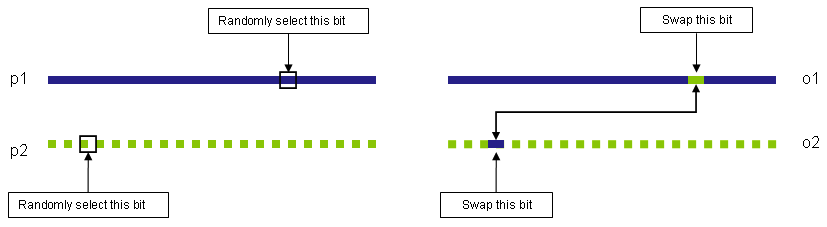}
\caption{1-Bit-Swap Operator}
\label{fig:fig1}
\end{figure}

This genetic recombination operator (see Figure \ref{fig:fig1}) was introduced in \cite{tersarkisov2010} and proved to work efficiently both alone and together with mainstream operators (crossovers and mutation). Its efficiency was mostly visible on functions like Rosenbrock, Ackley, Rastrigin and Royal Roads. We also tested its performance on OneMax specifically for this article.  
\subsection{Our findings}
The models we derive are complete, i.e. they are functions of just population size $\mu$, recombination pool $\lambda$ and length of the chromosome $n$, i.e. the actual parameters of EAs, though we make some weak assumptions about the pairing of parents.\\
\linebreak
We derive the expectation of convergence time for the population-based elitist EA with a recombination operator (1-Bit-Swap Operator) and mutation-based RLS. Our theoretical and computational findings confirm that for OneMax the benefit of population is unclear, i.e. its effect is not always positive. For Royal Roads it is always positive. This problem-specific issue was noticed before (see e.g. Figures 4 and 5 in \cite{heyao02}).\\ 
\linebreak
We use two distributions of elite species in the population: Uniform($\frac{1}{\mu})$ and Poisson(1). Since the expressions for the expected first hitting time of algorithms $\mathbf{E}\tau$ we have found are quite complicated, we do the computational estimation and find some asymptotic results as well.   
\begin{table}
\centering
\begin{tabular}{|r|l|}
\hline
\multicolumn{2}{|c|}{Algorithm 1:($\mu+\lambda$)EA$_{1BS}$}\\
\hline
1&Initialize population size $\mu$\\
& repeat until condition fulfilled:\\
2a& $\quad$ select $\lambda$ species from the population using\\
& $\quad$Tournament selection\\
2b& $\quad$ apply 1BS operator to each pair in the\\
& $\quad$ recombination pool\\ 
2c& $\quad$ keep $\alpha$ best species in the population, \\
& $\quad$ replace the rest with the best species\\
& $\quad$ from the pool\\
\hline
\end{tabular}
\linebreak
\begin{tabular}{|r|l|}
\hline
\multicolumn{2}{|c|}{Algorithm 2:$(\mu+\lambda)$RLS}\\
\hline
1&Initialize population size $\mu$\\
& repeat until condition fulfilled:\\
2a& $\quad$ select $\lambda$ species from the population using\\
& $\quad$ Tournament selection\\
2b& $\quad$ flip exactly one bit per chromosome\\
2c& $\quad$ keep $\alpha$ best species in the population, \\
& $\quad$ replace the rest with the best species \\
& $\quad$ from the pool\\
\hline
\end{tabular}
\caption{Pseudocode for the algorithms analyzed in this article}
\end{table}
\paragraph{Tournament Selection Procedure} We use this selection because it is fairly straightforward in implementation and analysis.
\begin{itemize}
\item Select two species $x_{i},x_{j}$ uniformly at random\\
\item if $f(x_i)=f(x_j)$, either $x_i$ or $x_j$ enters the pool at random \\ 
\item else the species with better fitness enters the pool 
\end{itemize}
\section{\uppercase{Analysis of Algorithm 1 on OneMax problem}}
\noindent We start with Uniform distribution of elite species with parameter $\frac{1}{\mu}$, which gives the lower bound on convergence time, which is due to the assumption on the number of elite species needed for the evolution.\\
\linebreak
The probability of selecting an elite pair is 
{\smalll
\begin{align}
P_{sel}&= \Big(\frac{\alpha}{\mu}\Big)^4+4\Big(\frac{\alpha}{\mu}\Big)^3\frac{\beta}{\mu}+4\Big(\frac{\alpha}{\mu}\Big)^2\Big(\frac{\beta}{\mu}\Big)^2 \nonumber \\
&=\frac{\alpha^2(\alpha+2\beta)^2}{\mu^4}=\frac{\alpha^2(\alpha+2(\mu-\alpha))^2}{\mu^4}        
\label{eq:prob_select}
\end{align}
}
Since we restrict the analysis only to elite pairs, the probability of evolution (generation of a better offspring as a result of 1-Bit-Swap) is 
{\smalll
\begin{equation}
P_{swap}=\frac{1}{2}-2\Big(\frac{k}{n}\Big)^2
\end{equation} 
\label{eq:swap_1bs_onemax}
}
where {\smalll $k=0:\frac{n}{2}-1$}, which is due to the assumption that at the start of the algorithm {\smalll $f(\alpha)=\frac{n}{2}$}. 
\subsection{Uniform distribution of elite species}
We are deriving an upper bound on the probability (and, therefore, lower bound on the expectation of the first hitting time). We are interested in the probability of evolving at least 1 new elite species next generation, i.e. of at least 1 successful swap.  
\linebreak
\begin{equation*}
P(\textnormal{ at least 1 new elite species in the population at } t+1)\\
\end{equation*}    
\begin{equation*}
=1-P(\textnormal{no new elite species in the population at } t+1)
\end{equation*}
\linebreak
We define $G_0$ to be the event that no new species evolves over 1 particular generation. The number of elite pairs $H_j$ in the population varies from 0 to $\frac{\lambda}{2}$, and elite species $\alpha$ in the population from 1 to $\mu$. In this regard, probability to select a number of pairs given $\alpha$ elite species in the population is $P_{sel \alpha}$. By the Law of total probability,  
{\smalll
\begin{align}
P(G_0)&=\sum_{j=0}^{\frac{\lambda}{2}}P(G_0|H_j)P(H_j) \nonumber \\
&=\sum_{j=0}^{\frac{\lambda}{2}}P(G_0|H_j)\sum_{\alpha=1}^{\mu}P(H_j|\alpha)P(\alpha)
\label{eq:ltp1}
\end{align}
}
We assume Uniform probability of each number of elite species in the population: $P(\alpha)=\frac{1}{\mu}$.
{\smalll
\begin{align}
&P(G_0|H_0)P(H_0)=P(G_0|H_0)\sum_{\alpha=1}^{\mu}P(H_0|\alpha)P(\alpha) \nonumber \\
&=\sum_{\alpha=1}^{\mu}P(H_0|\alpha)P(\alpha) \nonumber \\
&=\Big(1-P_{sel1}\Big)^{\frac{\lambda}{2}}+ \Big(1- P_{sel2} \Big)^{\frac{\lambda}{2}}+\ldots+\Big(1- P_{sel \mu} \Big)^{\frac{\lambda}{2}} \nonumber \\
&=\sum_{\alpha=1}^{\mu}\Big(1- P_{sel \alpha}\Big)^{\frac{\lambda}{2}}
\end{align} 
}
to get 1 elite pair:
{\smalll
\begin{equation*}
P(G_0|H_1)P(H_1)=\Big(\frac{1}{2}+2\Big(\frac{k}{n}\Big)^2\Big)\frac{1}{\mu}\sum_{\alpha=1}^{\mu}P(H_1|\alpha)
\end{equation*}}
and 
{\smalll
\begin{equation*}
\sum_{\alpha=1}^{\mu}P(H_1|\alpha)=\binom{\frac{\lambda}{2}}{1}\sum_{\alpha=1}^{\mu}  P_{sel \alpha} \Big(1- P_{sel \alpha} \Big)^{\frac{\lambda}{2}-1}
\end{equation*}}
therefore, the probability of failure given 1 elite pair given $k$ improvements so far is
{\smalll
\begin{align}
&P(G_{0k}|H_1)P(H_1)=\Big(\frac{1}{2}+2\Big(\frac{k}{n}\Big)^2\Big)\binom{\frac{\lambda}{2}}{1}\frac{1}{\mu}\sum_{\alpha=1}^{\mu} P_{sel \alpha}\Big(1-P_{sel \alpha})^{\frac{\lambda}{2}-1}
\end{align}}
For cases $\{H_j:2 \leq j \leq \frac{\lambda}{2}\}$ the logic is similar, so the full expression for the probability of failure is   
{\smalll
\begin{align*}
&P(G_{0k})=\sum_{j=0}^{\frac{\lambda}{2}}P(G_0|H_j)P(H_j)\\
&=\frac{1}{\mu}\sum_{j=0}^{\frac{\lambda}{2}} \Big(\frac{1}{2}+2\Big(\frac{k}{n}\Big)^2\Big)^j\binom{\frac{\lambda}{2}}{j}\sum_{\alpha=1}^{\mu} P_{sel \alpha}^j\Big(1-P_{sel \alpha} \Big)^{\frac{\lambda}{2}-j} \\
&=\frac{1}{\mu}\sum_{j=0}^{\frac{\lambda}{2}} \left(\frac{1}{2}+2\left(\frac{k}{n}\right)^2\right)^j\binom{\frac{\lambda}{2}}{j} \cdot\\
&\cdot \sum_{\alpha=1}^{\mu} \left(\frac{\alpha^2(\alpha+2(\mu-\alpha))^2}{\mu^4}\right)^j\left(1-\frac{\alpha^2(\alpha+2(\mu-\alpha))^2}{\mu^4}\right)^{\frac{\lambda}{2}-j}
\end{align*}
}
Interchanging the sums and using the standard binomial expansion $(s+t)^n=\sum_{k=0}^{n}\binom{n}{k}s^kt^{n-k}$
{\smalll
\begin{align*}
&P(G_{0k})=\frac{1}{\mu}\sum_{\alpha=1}^{\mu}\sum_{j=0}^{\frac{\lambda}{2}}\binom{\frac{\lambda}{2}}{j} \Big\{\Big(\frac{1}{2}+2\Big(\frac{k}{n}\Big)^2\Big)\left(\frac{\alpha^2(\alpha+2(\mu-\alpha))^2}{\mu^4}\right)\Big\}^j \\
&\Big\{\left(1-\frac{\alpha^2(\alpha+2(\mu-\alpha))^2}{\mu^4}\right)\Big\}^{\frac{\lambda}{2}-j} \\
&=\frac{1}{\mu^{2 \lambda+1}}\sum_{\alpha=1}^{\mu}\Big\{\mu^4-\Big(\frac{1}{2}-2\Big(\frac{k}{n}\Big)^2\Big)(\alpha(\alpha+2(\mu-\alpha)))^2 \Big\}^{\frac{\lambda}{2}}
\end{align*}
}
The probability of evolution (obtaining a better species) is therefore
{\smalll
\begin{align}
&P(G_k)=1-P(G_{0k}) \nonumber \\
&=1-\frac{1}{\mu^{2 \lambda+1}}\sum_{\alpha=1}^{\mu}\Big\{\mu^4-\Big(\frac{1}{2}-2\Big(\frac{k}{n}\Big)^2\Big)(\alpha(\alpha+2(\mu-\alpha)))^2 \Big\}^{\frac{\lambda}{2}}
\end{align}
}
for each $0 \leq k \leq \frac{n}{2}-1$ we have 
{\smalll
\begin{align*}
\mathbf{E}T_{k}&=\frac{1}{1-\frac{1}{\mu^{2 \lambda+1}}\sum_{\alpha=1}^{\mu}\Big\{\mu^4-\Big(\frac{1}{2}-2\Big(\frac{k}{n}\Big)^2\Big)(\alpha(\alpha+2(\mu-\alpha)))^2 \Big\}^{\frac{\lambda}{2}}}\\
&=\frac{\mu^{2 \lambda+1}}{\mu^{2 \lambda+1}-\sum_{\alpha=1}^{\mu}\Big\{\mu^4-\Big(\frac{1}{2}-2\Big(\frac{k}{n}\Big)^2\Big)(\alpha(\alpha+2(\mu-\alpha)))^2 \Big\}^{\frac{\lambda}{2}}}
\end{align*}
}
and therefore the expected first hitting time for the algorithm is 
{\smalll
\begin{align}
&\mathbf{E}\tau_{(\mu+\lambda)EA_{1BS}}=\sum_{k=0}^{\frac{n}{2}-1}\mathbf{E}T_k \nonumber \\
&=\mu^{2 \lambda+1}\sum_{k=0}^{\frac{n}{2}-1}\frac{1}{\mu^{2 \lambda+1}-\sum_{\alpha=1}^{\mu}\Big\{\mu^4-\Big(\frac{1}{2}-2\Big(\frac{k}{n}\Big)^2 \Big)(\alpha(\alpha+2(\mu-\alpha)))^2 \Big\}^{\frac{\lambda}{2}}} \nonumber \\
&=\mu^{2 \lambda+1} \phi(\lambda,\mu,n)
\label{eq:main_eq1}
\end{align}
}

\noindent Despite having two sums without closed forms, the convergence rate of this algorithms depends only on the size of the population, recombination pool and the length of the string, that is, the real-life parameters of EA. Therefore, the model is complete.\\
\section{\uppercase{Analysis of Algorithm 2 solving OneMax}}
\label{sec:pop_rls_onemax}
\noindent For comparison, we derive $\mathbf{E}\tau$ for $(\mu+\lambda)$EA$_{RLS}$ using similar approach (Law of total probability + sum of Geometric RVs). Changes apply mostly to the selection probability, as we have no pairs to form:
{\smalll \begin{equation}
P_{sel}=\Big(\frac{\alpha}{\mu}\Big)^2+\frac{2 \alpha \beta}{\mu^2}=\frac{\alpha(2\mu-\alpha)}{\mu^2}
\end{equation}}
\subsection{Uniform distribution of elite species}
We use the same assumptions of uniform distribution of elite species in the population as with the  $(\mu+\lambda)$EA$_{1BS}$.\\
\linebreak 
Failure event $G_0$ is defined in the same way: no successful flips in the recombination pool, so the probability thereof is defined in a similar way to the one in Equation \ref{eq:ltp1}.
{\smalll \begin{equation}
P(G_0)=\sum_{j=0}^{\lambda}P(G_0|H_j)\sum_{\alpha=1}^{\mu}P(H_j|\alpha)P(\alpha)
\label{eq:ltp2}
\end{equation}}
Only in this case $j$ is the number of elite parents in the pool and goes from 0 to $\lambda$.
{\smalll \begin{equation*}
P(H_j|\alpha)P(\alpha)=\binom{\lambda}{j}p_{sel}^j(1-p_{sel})^{\lambda-j}
\end{equation*}}
and therefore (using the same idea with the binomial expansion)
{\smalll \begin{align}
&P(G_{0k})=\frac{1}{\mu}\sum_{j=0}^{\lambda}\Big(\frac{k}{n}\Big)^j \binom{\lambda}{j}\sum_{\alpha=1}^{\mu}P_{sel}^j(1-P_{sel})^{\lambda-j} \nonumber \\
&=\frac{1}{\mu}\sum_{\alpha=1}^{\mu} \sum_{j=0}^{\lambda}\binom{\lambda}{j}\Big(\frac{k}{n}P_{sel}\Big)^j(1-P_{sel})^{\lambda-j}  \nonumber \\
&=\frac{1}{\mu^{2 \lambda+1}}\sum_{\alpha=1}^{\mu}[\mu^2-\alpha(2 \mu-\alpha)(1-\frac{k}{n})]^\lambda
\end{align}}
Unfortunately, the closed expression for this sum exists only for specific values of $\lambda$, so we have to keep it this way and later obtain the results computationally.
{\smalll \begin{equation*}
P(G_k)=1-P(G_{0k})=1-\frac{1}{\mu^{2 \lambda+1}}\sum_{\alpha=1}^{\mu}[\mu^2-\alpha(2 \mu-\alpha)(1-\frac{k}{n})]^\lambda
\end{equation*}}
So the expected optimization time of the algorithm is 
{\smalll
\begin{align}
&\mathbf{E}\tau_{(\mu+\lambda)RLS}=\sum_{k=\frac{n}{2}}^{n-1}\mathbf{E}T_k =\sum_{k=\frac{n}{2}}^{n-1}\frac{1}{P(G_k)} \nonumber \\
&=\mu^{2\lambda+1}\sum_{k=\frac{n}{2}}^{n-1}\frac{1}{\mu^{2 \lambda+1}-\sum_{\alpha=1}^{\mu}(\mu^2-\alpha(2\mu-\alpha)(1-\frac{k}{n}))^{\lambda}}
\label{eq:rls_1max_unif}
\end{align}}

\noindent As the case is with $(\mu+\lambda)$EA$_{1BS}$, this is a somewhat optimistic estimate, since it assigns fairly high probabilities to high proportions of elite species in the population. This is confirmed by numerical estimates. 
\section{\uppercase{Analysis of Algorithm 1 On Royal Roads Function}}
\label{sec:rr}
\noindent We use the setup for RR problem along the lines of \cite{mitchell96} (referred to as $R_1$ in the book).The chromosome of length $n$ is split into $K$ blocks, each of length $M$. The fitness of each block is 0 if there are any 0s in the block, and M if all of the bits in it have value 1. The fitness of the chromosome is the sum of the value of the blocks, so it can take values $0, K, 2K, \ldots MK$. We index the blocks using index $\kappa$. Originally this problem was designed to test EA's capacity for recombining building blocks compared to other heuristics (for details see \cite{mitchell96}).\\
\linebreak
Additionally, we introduce an auxiliary function used to measure progress between improvements in the fitness (the idea is similar to that in, e.g. \cite{doerr103,heyao04}), which in our case is $V_{\kappa}=\textnormal{OneMax}(s_{\kappa})$ since both functions achieve the global optimum at $s_k=M$ and $\max f(s)=V_s=\sum V_{\kappa}=n$.\\
\linebreak
There is an important difference from the standard OneMax problem: unlike it, when parents exchange genetic information, it doesn't matter where the information comes from (which segment of the parent), but it matters where it is inserted, because it may mean that the fitness of the recipient segment has reached $M$, and therefore the fitness of the whole parent increased by the same value.\\
\linebreak
The second important observation is that of all segments in the chromosome there is one, which evolves first, denote it $\kappa_1$ (this is only possible due to the parameters of the EA in discussion). This means that segments evolve in a sequence: $\kappa_1,\kappa_2,\ldots,\kappa_k$.\\
\linebreak
We pessimistically assume that the best auxiliary function value in the first generation is $\frac{n}{2}$ and fitness function is 0. We also assume that the starting value in each bin $\kappa$ is $\frac{M}{2}$. In the same way as with OneMax, we make assumptions about the distribution of elite species in the population, rather than their exact or approximate number.\\
\linebreak
We start with introducing the probability of failure:
{\smalll
\begin{equation*}
P(G_{0})=\sum_{j=0}^{\frac{\lambda}{2}}P(G_{0}|H_j)\sum_{\alpha=1}^{\mu}P(H_j|\alpha)P(\alpha)
\end{equation*}}
where all variables are the same as in $(\mu+\lambda)$EA$_{1BS}$ solving OneMax: $H_j$ is j'th elite pair in the recombination pool $\lambda$, $\alpha$ is the number of elite species in the population $\mu$ with both highest fitness and auxiliary function values. The selection function is the same as Equation \ref{eq:prob_select}:
{\smalll
\begin{equation*}
P_{sel}(\alpha)=\frac{\alpha^2(\alpha+2(\mu-\alpha))^2}{\mu^4}    
\end{equation*}}
The successful event is defined as evolution of at least one more elite species in the population. The number of bits equal to 0 left to swap/flip in a segment we use $l$. So now the probability of successful swap in Equation \ref{eq:swap_1bs_onemax} becomes 
{\smalll \begin{equation}
P_{swap}=\frac{2\Big(\frac{M}{2}-l\Big)\Big(\frac{n}{2}+\frac{kM}{2}+l\Big)}{n^2} =\frac{(M-2l)(n+kM+2l)}{2n^2}
\label{eq:prob_swap_rr}
\end{equation}}
The auxiliary function for each bin $\kappa, V_ {\kappa}$ lies between $0$ and $\frac{M}{2}-1$ and k between $0$ and $K-1$, where K is the total number of bins $\kappa$ to fill. The probability of failure is
{\smalll \begin{align*}
&P_{F}=1-P_{swap}=1-\frac{(M-2l)(n+kM+2l)}{2n^2} \nonumber \\
&=\frac{2n^2-(M-2l)(n+kM+2l)}{2n^2}
\end{align*}}
The probability to fail to improve a bit in a bin given $l$ improvements so far is 
{\small
\begin{align}
&P(G_{0 l})=\frac{1}{\mu}\sum_{j=0}^{\frac{\lambda}{2}}\Big(\frac{2n^2-(M-2l)(n+kM+2l)}{2n^2}\Big)^j \binom{\frac{\lambda}{2}}{j} \nonumber \\
&\sum_{\alpha=1}^{\mu}\Big(\frac{(\alpha(\alpha+2 \mu(\mu-\alpha)))}{\mu^2}\Big)^j \Big(1-\frac{(\alpha(\alpha+2 \mu(\mu-\alpha)))}{\mu^2}\Big)^{\frac{\lambda}{2}-j} \nonumber \\
&=\frac{1}{\mu}\sum_{j=0}^{\frac{\lambda}{2}}P_{F}^{j}\binom{\frac{\lambda}{2}}{j}\sum_{\alpha=1}^{\mu}(P_{sel}(\alpha))^j(1-P_{sel}(\alpha))^{\frac{\lambda}{2}-j} \nonumber \\
&=\frac{1}{\mu}\sum_{\alpha=1}^{\mu}(1-P_{sel}(\alpha)P_{swap})^{\frac{\lambda}{2}}
\end{align}
}
Therefore,
{\smalll \begin{equation*}
P(G_l)=1-P(G_{0l})=1-\frac{1}{\mu}\sum_{\alpha=1}^{\mu}(1-P_{sel}(\alpha)P_{swap})^{\frac{\lambda}{2}}
\end{equation*}}
Expected time until improving the auxiliary function of a bin $\kappa$ is 
{\smalll \begin{equation}
\mathbf{E}T_{\kappa}=\sum_{l=0}^{\frac{M}{2}-1}\frac{1}{P(G_l)}
\end{equation}}
and, finally,summing over all $k$ from 0 to $K-1$ we obtain (since G depends on both $l$ and $k$)
{\smalll \begin{align}
&\mathbf{E}\tau_{(\mu+\lambda)EA_{1BS}}=\sum_{k=0}^{K-1}\sum_{l=0}^{\frac{M}{2}-1}\frac{1}{P(G_{l,k})} \nonumber \\
&=\mu^{2 \lambda+1}\sum_{k=0}^{K-1}\sum_{l=0}^{\frac{M}{2}-1}\frac{1}{\mu^{2 \lambda+1}-\sum_{\alpha=1}^{\mu}(\mu^4-(\alpha(\alpha+2(\mu-\alpha)))^2 P_{swap})^{\frac{\lambda}{2}}}
\label{eq:rr_1bs_unif}
\end{align}}
\section{\uppercase{Analysis of Algorithm 2 on Royal Roads Function}}
Just as is the case with OneMax, we present the results for population-based RLS on Royal Roads. This model is a bit simpler since we do not have to pair the parents, and the selection is just 
{\smalll \begin{equation*}
P(\alpha)_{sel}=\Big(\frac{\alpha}{\mu}\Big)^2+\frac{2 \alpha \beta}{\mu^2}=\frac{\alpha(2 \mu-\alpha)}{\mu^2}
\end{equation*}}
Instead of Uniform distribution of elite species in the population, we try Poisson distribution with parameter 1, and normalizing constant
{\smalll \begin{equation}
c(\mu)=\frac{e}{\sum_{\alpha=1}^{\mu}\frac{1}{\alpha!}}=\frac{e \Gamma(\mu+1)}{e \Gamma(\mu+1,1)-\Gamma(\mu+1)}
\label{eq:norm_const2}
\end{equation}}
since {\smalll $\sum_{k=0}^{n}\frac{x^k}{k!}=\frac{e^{\lambda}\Gamma(n+1,\lambda)}{\Gamma(n+1)}$} where {\smalll $\Gamma(n+1,\lambda)$} is incomplete Gamma function. The sizes of populations used in the computational experiments, the values of the normalizing constant are set in Table \ref{tab:nc2}.
{\smalll
\begin{table}
\centering
\begin{tabular}{|c|c|}
\hline
$\mu$&$c(\mu)$\\
\hline
4&$\frac{24 e}{41}$\\
10&$\approx 1.58198$\\
20&$\approx 1.58198$\\
30&$\approx 1.58198$\\
\hline
\end{tabular}
\caption{Values of the normalizing constant, Equation \ref{eq:norm_const2}}
\label{tab:nc2}
\end{table}
}\\
\linebreak
The flip probability is just 
{\smalll \begin{equation}
P_{flip}=\frac{M-2l}{n}
\end{equation}
Probability of failure given $l$ successful flips so far is
\begin{align*}
&P(G_{0l})=\frac{c(\mu)}{e}\sum_{j=0}^{\lambda}\Big(1-\frac{M-2l}{2n}\Big)^j\binom{\lambda}{j} \nonumber \\
&\cdot \sum_{\alpha=1}^{\mu}\Big(\frac{\alpha(2 \mu-\alpha)}{\mu^2}\Big)^j \Big(1-\frac{\alpha(2 \mu-\alpha)}{\mu^2}\Big)^{\lambda-j} \frac{1}{\alpha!} \nonumber \\
&=\frac{c(\mu)}{e}\sum_{\alpha=1}^{\mu}\frac{1}{\alpha!}\Big[1-\frac{\alpha(2 \mu -\alpha)}{\mu^2}\Big(\frac{M-2l}{2n}\Big)\Big]^\lambda \\
&=\frac{c(\mu)}{e \mu^{2 \lambda}}\sum_{\alpha=1}^{\mu}\frac{1}{\alpha!}\Big[\mu^2-\alpha(2 \mu-\alpha)\Big(\frac{M-2l}{2n}\Big)\Big]^{\lambda}
\end{align*}}
Therefore, the probability of success is 
{\smalll \begin{equation*}
P(G_l)=1-P(G_{0l})
\end{equation*}}
and the expected time to fill the first bin $\kappa_1$ is therefore 
{\smalll
\begin{align*}
&\mathbf{E}T_{\kappa 1}=\sum_{l=0}^{\frac{M}{2}-1}\frac{1}{1-\frac{c(\mu)}{e\mu^{2 \lambda}}\sum_{\alpha=1}^{\mu}\frac{1}{\alpha!}\Big[\mu^2-\alpha(2 \mu-\alpha)\Big(\frac{M-2l}{2n}\Big)\Big]^{\lambda}} \\
&=e \mu^{2 \lambda}\sum_{l=0}^{\frac{M}{2}-1}\frac{1}{e\mu^{2 \lambda}-c(\mu)\sum_{\alpha=1}^{\mu}\frac{1}{\alpha!}\Big[\mu^2-\alpha(2 \mu-\alpha)\Big(\frac{M-2l}{2n}\Big)\Big]^{\lambda}}
\end{align*}
}
\noindent Since we have $k$ such bins and the probability of successful sampling does not depend on the number of 1's in the parent (unlike $(\mu+\lambda)$EA$_{1BS}$), we obtain the expected first hitting time for the algorithm on RR:
{\smalll
\begin{align}
&\mathbf{E}\tau_{(\mu+\lambda)RLS}=e\mu^{2 \lambda} \nonumber \\
&\sum_{k=0}^{K-1}\sum_{l=0}^{\frac{M}{2}-1}\frac{1}{e\mu^{2 \lambda}-c(\mu)\sum_{\alpha=1}^{\mu}\frac{1}{\alpha!}\Big[\mu^2-\alpha(2 \mu-\alpha)\Big(\frac{M-2l}{2n}\Big)\Big]^{\lambda}} \nonumber \\
&=K e \mu^{2 \lambda} \sum_{l=0}^{\frac{M}{2}-1}\frac{1}{e\mu^{2 \lambda}-c(\mu)\sum_{\alpha=1}^{\mu}\frac{1}{\alpha!}\Big[\mu^2-\alpha(2 \mu-\alpha)\Big(\frac{M-2l}{2n}\Big)\Big]^{\lambda}}
\label{eq:rls_rr_poisson}
\end{align}
}
\smalll{
\begin{table*}[t]
\centering
\begin{tabular}{|c|c|c|c|c|c|c|}
\hline
$n$&$\mu$&$\lambda$&$\mathbf{E}\tau^{R}_{Algorithm 1}$&$\tilde{\tau}_{Algorithm 1}$&$\mathbf{E}\tau^{R}_{Algorithm 2}$&$\tilde{\tau}_{Algorithm 2}$\\
\hline
\multirow{5}{*}{50}&1&2&112.9801&113.14&190.7979&192.12\\
&2&2&144.6145&218.12&116.2812&184.6\\
&4&4&94.7621&145.4&73.124&193.54\\
&8&8&62.5691&121.86&48.596&184.52\\
&10&10&55.4784&116.98&43.488&197.92\\
\hline
\multirow{5}{*}{100}&1&2&259.8688&265.84&449.9205&418.8\\
&2&2&332.6321&455.9048&271.56&393.39\\
&4&4&215.2445&314.1&168.03&410.39\\
&8&8&139.2885&267.72&108.826&420.416\\
&10&10&122.4884&266.9&96.3978&405.22\\
\hline
\multirow{5}{*}{1000}&1&2&3743.2354&3682.6&6792.8&6715\\
&2&2&4791.3413&7072.7&4025.9876&6630\\
&4&4&3021.5468&4574.3&2413.0033&6990\\
&8&8&1872.3595&3866.5&1481.761&7016\\
&10&10&1616.4433&3807&1283.5502&6834\\
\hline
\end{tabular}
\caption{Theoretical and computational bounds for OneMax test function with the assumption of Uniform distributions of elite species in the population (Equations \ref{eq:main_eq1} and \ref{eq:rls_1max_unif})}
\end{table*}}
\section{\uppercase{Computational results}}
\noindent Since the expressions derived in this article do not have a closed form, we find them computationally. To test our results, we run each algorithm with parameter set ($\mu,\lambda,n$) with $\mu=\lambda$ almost always for 50 independent runs, each run was 2000 generations long. The average of optimization time is denoted $\tilde{\tau}$. Probability distribution used for each model follow standard notation in Probability theory: $R$ for Uniform and $P$ for Poisson.\\
\linebreak
In general, results for Algorithm 1 tend to be better than for Algorithm 2 and for OneMax sharper than for RR. As we mentioned already, this is due to different patterns of dynamics of elite species and has to be investigated further. Apparently for both algorithms solving RR both Uniform$(\frac{1}{\mu})$ and Poisson(1) distributions give a fairly rough approximation that we can improve both statically (using other parameters) and dynamically (modeling change in the number of elite species).\\
\linebreak
The other important result is that the increase of population size for both algorithms solving OneMax problem does not necessarily result in the improvement in performance, which we showed both theoretically and numerically. For RR the situation is much more clear: increase in the population always brings about the improvement in performance. Both models confirm this quite consistently.  
\begin{center}
{\smalll
\begin{table*}[t]
\centering
\begin{tabular}{|c|c|c|c|c|c|c|c|c|}
\hline
$n$&K&M&$\mu$&$\lambda$&$\mathbf{E}\tau^{R}_{Algorithm 1}$&$\tilde{\tau}_{Algorithm 1}$&$\mathbf{E}\tau^{P}_{Algorithm 2}$&$\tilde{\tau}_{Algorithm 2}$\\
\hline
\multirow{4}{*}{32}&\multirow{4}{*}{4}&\multirow{4}{*}{8}&4&4&145&315.3077&64.8084&672.25\\
&&&10&10&72.4&268.2195&58.124& 504.625\\
&&&20&20&44.2&192.2917&56.175&334.125\\
&&&30&30&34.5&173.5625&55.1&221 \\
\hline
\multirow{4}{*}{64}&\multirow{4}{*}{8}&\multirow{4}{*}{8}&4&4&570.625&612.46&249.959&-\\
&&&10&10&279.88&497.93&222.565&820.6667\\
&&&20&20&153.46&454.4681&212.452&715.92\\
&&&30&30&112.297&372.04&209.373&663.6744\\
\hline
\multirow{4}{*}{128}&\multirow{4}{*}{16}&\multirow{4}{*}{8}&4&4&2264.36&1365&1021&-\\
&&&10&10&1048&1239&940.999&-\\
&&&20&20&570.44&1091.5&887.396&1612\\
&&&30&30&401.99&949.4&871.1131&1505\\
\hline
\end{tabular}
\caption{Theoretical and computational bounds for Royal Roads test function with the assumptions of Uniform and Poisson distributions of elite species in the population (Equations \ref{eq:rr_1bs_unif} and \ref{eq:rls_rr_poisson}).}
\end{table*}}
\end{center}
\section{\uppercase{conclusions and future work}}
\noindent We have derived expected running time for EAs based on two different genetic operators on two relatively simple fitness functions. In the future there are two extensions that we particularly plan to focus on:
\paragraph{Approximate results for Equations \ref{eq:main_eq1}, \ref{eq:rls_1max_unif}, \ref{eq:rr_1bs_unif}, \ref{eq:rls_rr_poisson}} This is the most obvious of developments. Although these equations give good estimates for optimization time, and we have found some asymptotic lower bounds, it is desirable to find sharper bounds in the form {\smalll $O(g(\mu, \lambda, n))$}. A big problem here are the complicated expressions involving sums. 
\paragraph{Evolution of elite species.} This area has seen little focus in EA community, and we are keen to develop a dynamic model for evolution of species. In this article the model is static, i.e. distribution of elite species in the population is fixed (Uniform or Poisson). As a result, some bounds, especially for RR, seem to be quite loose. If instead of assuming a probability distribution of elite species with fixed parameters we study the convergence of the distribution, we can derive sharper bounds on optimization time.\\
\bibliographystyle{apalike}
{\small
\bibliography{mybib6}}
\end{document}